\documentclass{article} 
\usepackage{iclr2026_conference,times}

\usepackage{hyperref}
\usepackage{url}
\usepackage{lineno}

\usepackage[table]{xcolor}
\usepackage{microtype}
\usepackage{tikz}
\usetikzlibrary{shadows}
\usetikzlibrary{arrows}
\usetikzlibrary{decorations.pathreplacing}
\usepackage{multirow}
\usepackage{subcaption}
\usepackage{amssymb}
\usepackage{amsmath} 
\usepackage{todonotes}
\usepackage{float}
\usepackage{placeins}
\usepackage{color,soul}
\usepackage{booktabs}
\usepackage{wrapfig}
\usepackage{enumitem}

\title{MaxPoolBERT: Enhancing BERT Classification\\ via Layer- and Token-Wise Aggregation}

\author{Maike Behrendt, Stefan Sylvius Wagner \\
Heinrich Heine University Düsseldorf\\
Lamarr Institute for Machine Learning and AI \\
\small{\texttt{\{maike.behrendt, stefan.wagner\}@hhu.de}} \\
\And
Stefan Harmeling \\
Technical University Dortmund \\
Lamarr Institute for Machine Learning and AI \\
\small{\texttt{stefan.harmeling@tu-dortmund.de}} \\
}

\iclrfinalcopy 
\begin{document}

\maketitle

\begin{abstract}
The \texttt{[CLS]} token in BERT is commonly used as a fixed-length representation for classification tasks, yet prior work has shown that both other tokens and intermediate layers encode valuable contextual information. In this work, we study lightweight extensions to BERT that refine the \texttt{[CLS]} representation by aggregating information across layers and tokens. Specifically, we explore three modifications: (i) max-pooling the \texttt{[CLS]} token across multiple layers, (ii) enabling the \texttt{[CLS]} token to attend over the entire final layer using an additional multi-head attention (MHA) layer, and (iii) combining max-pooling across the full sequence with MHA. Our approach, called MaxPoolBERT, enhances BERT’s classification accuracy (especially on low-resource tasks) without requiring new pre-training or significantly increasing model size. Experiments on the GLUE benchmark show that MaxPoolBERT consistently achieves a better performance than the standard BERT base model on low resource tasks of the GLUE benchmark.
\end{abstract}

\section{Introduction} 

BERT (Bidirectional Encoder Representations from Transformers) \citep{devlin_etal_2019_bert}, is one of the best known Transformer-based \citep{NIPS2017_3f5ee243} language models. The core principle of BERT is the unsupervised pre-training approach on large corpora, enabling it to learn contextual word representations, which can then be used to solve various downstream tasks. Through fine-tuning, BERT adapts its representations to aggregate the most relevant information required for a given task. 

\begin{wrapfigure}{r}{0.5\textwidth}
    \centering
    \includegraphics[scale=0.27]{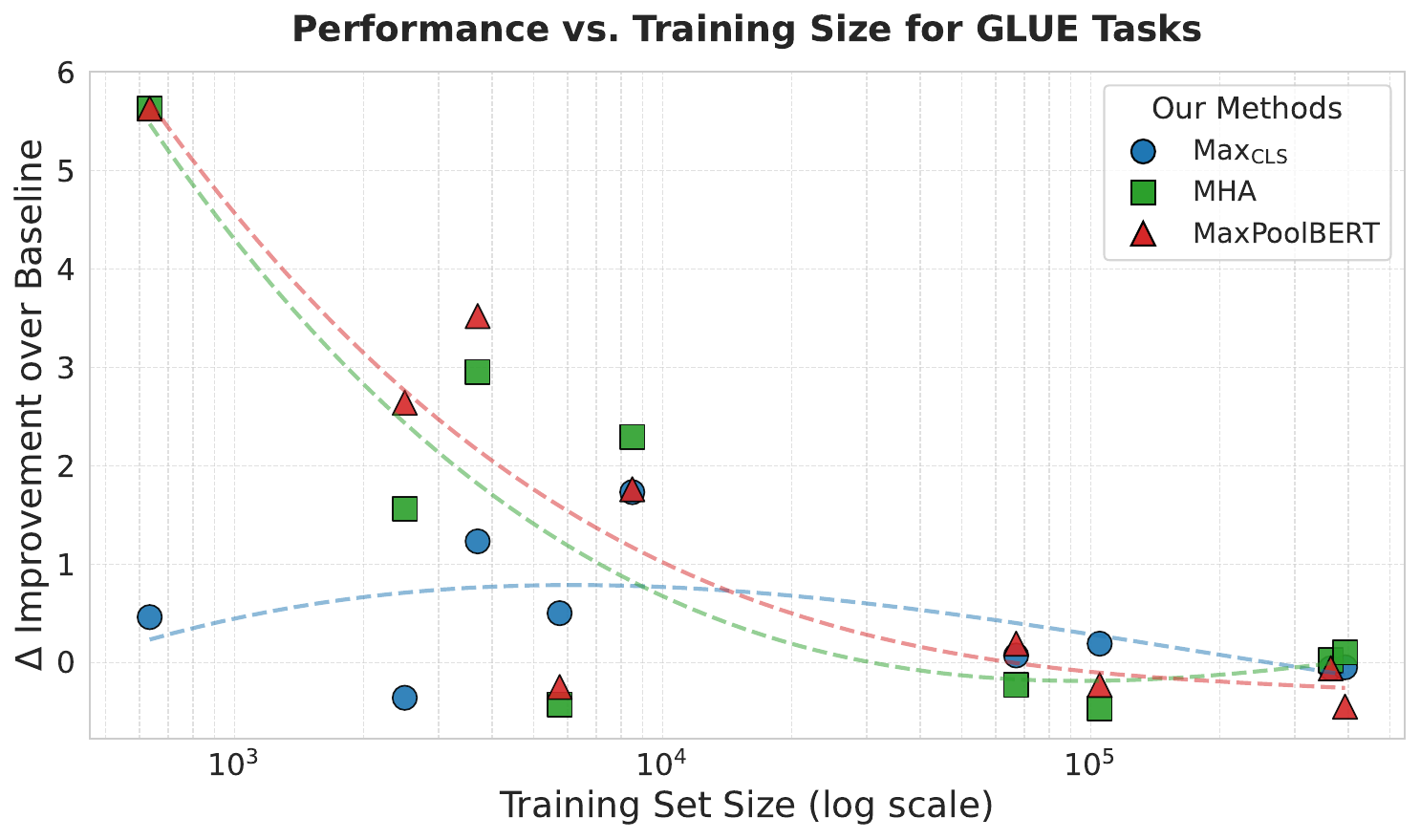}
    \caption{\textbf{MaxPoolBERT performs best on low-resource datasets.} We show that our methods, in particular MaxPoolBERT, provide significant improvements for smaller datasets indicating that the model learns a better representation during fine-tuning (top-left).}
    \label{fig:deltas}
\end{wrapfigure}

A key component of BERT's architecture is the classification token (abbreviated \texttt{[CLS]}), a special token that is prepended to every input sequence. During fine-tuning, the \texttt{[CLS]} token serves as the only input to the classification head, which generates predictions for the task at hand. Through self-attention, the \texttt{[CLS]} token is expected to capture the sentence-level information necessary for downstream tasks.   In this paper, we ask the question whether we can enrich the \texttt{[CLS]} token with information from the layers below the top level.

We know that the last layers of BERT change the most during fine-tuning and encode the most task-specific information \citep{rogers-etal-2020-primer}. This is why the \texttt{[CLS]} token embedding from the final layer is conventionally used for classification. However, assuming that only the \texttt{[CLS]} token retains meaningful sentence-level information is misleading. Prior studies have shown that all token embeddings in the final layer contain sentence-level information \citep{rogers-etal-2020-primer}, and that using different token positions for classification results in only minor differences in accuracy \citep{pmlr-v119-goyal20a}. \citet{pmlr-v119-goyal20a} also found that embedding vectors in the final layer exhibit high cosine similarity due to information mixing through self-attention.

Motivated by these findings, we explore incremental modifications to the BERT base architecture for sequence classification, aiming to enhance its performance on downstream tasks. We specifically focus on improving the informativeness of the \texttt{[CLS]} token by (i) incorporating more \emph{width} information of the whole sequence, and (ii) incorporating more \emph{depth} information from additional layers. In the end we find that a mixture of these approaches leads to the best results. 

\paragraph{Contributions.}
\begin{enumerate}
    \item We introduce MaxPoolBERT, an effective extension to BERT that enriches the \texttt{[CLS]} token representation using max-pooling and attention mechanisms across layers and tokens.
    \item We systematically evaluate three architectural variants that incorporate \emph{width} (token-level) and \emph{depth} (layer-level) information into the \texttt{[CLS]} embedding.
    \item We show that our proposed approach improves fine-tuning performance on 7 out of 9 GLUE tasks and achieves an average gain of 1.25 points over the BERT base baseline.
    \item We demonstrate that MaxPoolBERT is particularly effective in low-resource scenarios, providing improved stability and accuracy where training data is limited.
\end{enumerate}

All of our models will be made publicly available after the review process. 


\section{Related Work}
Much research has been done dedicated to improving and optimizing BERT's training process through architectural modifications and fine-tuning strategies. Below, we discuss advancements in fine-tuning stability, text representations, model enhancements, and training efficiency. Our work falls within the branch of research aimed at optimizing BERT's representation to enhance downstream classification results, with a particular focus on augmenting the informativeness of the \texttt{[CLS]} token.

\paragraph{Stabilized BERT Fine-Tuning.}
The pre-training and fine-tuning paradigm for language models such as BERT \citep{devlin_etal_2019_bert} has led to significant improvements across a wide range of NLP tasks while keeping computational costs manageable. However, fine-tuning remains unstable due to challenges like vanishing gradients \citep{mosbach2021on} and limited dataset sizes \citep{zhang2021revisiting}. Several studies have proposed techniques to address this instability.

\citet{zhang2021revisiting} explore re-initializing BERT layers before fine-tuning, demonstrating that retaining all pre-trained weights is not always beneficial for fine-tuning. They also show that extending fine-tuning beyond three epochs improves performance. 
\citet{hao-etal-2020-investigating} examine how fine-tuning affects BERT's attention, finding that higher layers change significantly while lower layers remain stable. They propose a noise regularization method to enhance stability.
\citet{mosbach2021on} identify high learning rates as a key issue that cause fine-tuning instability. They propose using small learning rates with bias correction and increasing training iterations until nearly zero training loss is achieved.  
\citet{hua-etal-2021-noise} introduce Layer-wise Noise Stability Regularization which further stabilizes fine-tuning through regularization.
\citet{10.1007/s11390-021-1119-0} propose self-ensemble and self-distillation mechanisms that enhance fine-tuning stability without requiring architectural changes or external data. 

Our method, while not explicitly targeting stability, contributes to more robust performance especially on low-resource tasks by enabling the \texttt{[CLS]} token to integrate a broader context via pooling and attention. We analyze fine-tuning stability of our variants in Section \ref{subsec:stability}.

\paragraph{Faster and More Efficient Training.}
In addition to stabilization, architectural enhancements have been introduced to boost BERT's efficiency and effectiveness.
\citet{pmlr-v119-goyal20a} propose to eliminate tokens after fine-tuning, to reduce the time of inference. They discovered that the token representations in the highest layer of BERT base carry similar information.

Recently, \citet{warner2024smarter} introduce ModernBERT, an updated version of BERT with an increased sequence length of 8192. ModernBERT incorporates architectural improvements such as GeGLU activations \citep{shazeer2020glu}, Flash Attention \citep{dao2022flashattention}, and RoPE embeddings \citep{10.1016/j.neucom.2023.127063}.

While other approaches improve the input embedding size of BERT \citep{nussbaum2024nomic} or refine the pre-training process for GPU's \citep{pmlr-v202-geiping23a, NEURIPS2023_095a6917, izsak-etal-2021-train}, our work specifically concentrates on optimizing the \texttt{[CLS]} token during fine-tuning, leveraging the information captured in BERT's layers after pre-training.

\paragraph{Improved BERT Fine-Tuning and Representation Learning.}
Lastly, several approaches refine BERT's classification capability through optimized fine-tuning strategies and enriched sentence representations - areas that align closely with our approach (see also \citet{app14198887} who provide a comprehensive survey of methods for extracting sentence-level embeddings from BERT).


\citet{toshniwal-etal-2020-cross} systematically compared different text span representations using BERT, and found that max-pooling performs quite well across tasks, though its effectiveness varies. \citet{bao-etal-2021-span-fine} construct sentence representations for classification by selecting meaningful n-grams and combining sub-tokens of a pre-trained BERT model into span representations using a max-pooling approach. In contrast, our method does not require span selection or input modification, and applies pooling and attention directly to hidden states during fine-tuning.
\citet{HU2024101646} introduce a flexible BERT architecture with dynamic width and depth that adapts the number of attention heads, hidden size, and number of layers at inference time using knowledge distillation. Our approach does not alter the base architecture, instead we enrich the fixed-size \texttt{[CLS]} embedding to boost classification performance.

\citet{chang-etal-2023-multi} introduce Multi-CLS BERT, a framework that modifies pre-training and adds multiple \texttt{[CLS]} tokens to the sequence for fine-tuning. We achieve comparable results on the GLUE benchmark without altering the pre-training setup.
\citet{chen2023improving} present HybridBERT, which incorporates a hybrid pooling network and drop masking during pre-training to accelerate training and improve downstream accuracy. While HybridBERT combines multiple pooling strategies (mean, max, and attention) to replace the \texttt{[CLS]} token, we retain the original \texttt{[CLS]} embedding and instead enrich it through architectural refinements such as an additional multi-head attention layer and optional sequence-wide pooling. This allows our method to be applied to any pre-trained BERT-like model without re-training, with particular benefits observed on tasks with limited training data.


Recently, \citet{galal2024rethinking} explore aggregation techniques such as mean pooling and self-attention on output embeddings for Arabic sentiment analysis. They show that freezing BERT during fine-tuning can boost performance. Our method can be combined with such techniques but focuses on improving the \texttt{[CLS]} pathway, especially under low-resource conditions.

Lastly, \citet{lehevcka2020adjusting} propose modifying BERT’s output pooling strategy to improve large-scale multi-label text classification. Specifically, they replace the \texttt{[CLS]} token with combined mean and max pooling over the final hidden states of all tokens, arguing that this captures richer semantic information for classification. While their method entirely discards the \texttt{[CLS]} embedding, our approach retains it and enhances its contextual richness by integrating sequence-wide information via additional architectural layers during fine-tuning.

\begin{figure*}[t]
    \centering
    \begin{subfigure}[t]{0.48\textwidth}
    \centering
	\begin{tikzpicture}[block/.style={rectangle, draw, line width=0.3mm, rounded corners, font=\ttfamily\scriptsize}]
		\node[block, minimum width=6.0cm, minimum height=0.9cm, line width=0.6mm, rounded corners=5mm, fill=cyan!10, drop shadow] (layer1) at (0,-3.3) {Layer 1};
		
		\node[minimum width=6.0cm, minimum height=0.3cm] (dots2) at (0,-2.5) {\dots};
		
		\node[block, minimum width=6.0cm, minimum height=0.9cm, line width=0.6mm, rounded corners=5mm, fill=cyan!10, drop shadow] (layern) at (0,-1.7) {Layer N};
		
		\node[minimum width=6.0cm, minimum height=0.3cm] (dots3) at (0,-0.6) {};
		
		\node[block, minimum width=6.0cm, minimum height=0.9cm, line width=0.6mm, rounded corners=5mm, fill=magenta!10, drop shadow] (clshead) at (0,-0.6) {Classification Head};
		
		\node[block, minimum width=1cm, minimum height=0.5cm, line width=0.3mm, align=center, fill=yellow!10] (cls) at (-2.3,-4.2) {[CLS]};
		\node[block, minimum width=1cm, minimum height=0.5cm, line width=0.3mm, align=center, fill=yellow!10] (tok1) at (-1.15,-4.2) {Tok1};
		\node[block, minimum width=1cm, minimum height=0.5cm, line width=0.3mm, align=center, fill=yellow!10] (tok2) at (0,-4.2) {Tok2};
		\node[minimum width=1cm, minimum height=0.5cm, line width=0.3mm, align=center] (dots) at (1.15,-4.2) {\dots};
		\node[block, minimum width=1cm, minimum height=0.5cm, line width=0.3mm, align=center, fill=yellow!10] (tokn) at (2.3,-4.2) {TokT};

        \node[block, minimum width=1cm, minimum height=0.5cm, line width=0.3mm, align=center, fill=green!10] (clslay1) at (-2.3,-3.3) {E\textsubscript{[CLS]}};

        \node[block, minimum width=1cm, minimum height=0.5cm, line width=0.3mm, align=center, fill=green!10] (clslayN) at (-2.3,-1.7) {E\textsubscript{[CLS]}};
  
		\draw[->, line width=0.3mm] (-2.3,-1.55) -- (-2.3,-1.05);
		
        \draw[->, line width=0.3mm] (2.3,-2.4) -- (2.3, -1.95);
		\draw[->, line width=0.3mm] (0,-2.4) -- (0, -1.95);
		\draw[->, line width=0.3mm] (-1.15,-2.4) -- (-1.15, -1.95);
		\draw[->, line width=0.3mm] (-2.3,-2.4) -- (-2.3,-1.95);

	    \draw[->, line width=0.3mm] (2.3,-3.05) -- (2.3, -2.6);
		\draw[->, line width=0.3mm] (0,-3.05) -- (0, -2.6);
		\draw[->, line width=0.3mm] (-1.15,-3.05) -- (-1.15, -2.6);
		\draw[->, line width=0.3mm] (-2.3,-3.05) -- (-2.3,-2.6);

		\draw[->, line width=0.3mm] (2.3,-3.95) -- (2.3, -3.5);
		\draw[->, line width=0.3mm] (0,-3.95) -- (0, -3.5);
		\draw[->, line width=0.3mm] (-1.15,-3.95) -- (-1.15, -3.5);
		\draw[->, line width=0.3mm] (-2.3,-3.95) -- (-2.3,-3.5);
		
	\end{tikzpicture}
    \caption{\textbf{Baseline.} Plain vanilla BERT for sequence classification, where the embedding of the \texttt{[CLS]} token of the final layer is used as input for the classification head.}
    \label{fig:bert}
\end{subfigure}\quad
\begin{subfigure}[t]{0.48\textwidth}
        \centering
	\begin{tikzpicture}[block/.style={rectangle, draw, line width=0.3mm, rounded corners, font=\ttfamily\scriptsize}]
		\node[block, minimum width=6.0cm, minimum height=0.9cm, line width=0.6mm, rounded corners=5mm, fill=cyan!10, drop shadow] (layer1) at (0,-3.3) {Layer 1}; 
		
		\node[minimum width=6.0cm, minimum height=0.3cm] (dots2) at (0,-2.5) {\dots};
		
		\node[block, minimum width=6.0cm, minimum height=0.9cm, line width=0.6mm, rounded corners=5mm, fill=cyan!10, drop shadow] (layern) at (0,-1.7) {Layer N}; 
		
		\node[minimum width=6.0cm, minimum height=0.3cm] (dots3) at (0,-0.6) {};
		
		\node[block, minimum width=6.0cm, minimum height=0.9cm, line width=0.6mm, rounded corners=5mm, fill=magenta!10, drop shadow] (attention) at (0,-0.6) {Max};

        \node[block, minimum width=6.0cm, minimum height=0.9cm, line width=0.6mm, rounded corners=5mm, fill=magenta!10, drop shadow] (attention) at (0,0.5) {Classification Head};

        \node[block, minimum width=1cm, minimum height=0.5cm, line width=0.3mm, align=center, fill=green!10] (clslayN) at (-2.3,-0.6) {E\textsubscript{[CLS]}};

        \node[block, minimum width=1cm, minimum height=0.5cm, line width=0.3mm, align=center, fill=green!10] (clslayN) at (-2.3,-1.7) {E\textsubscript{[CLS]}};

        \node[block, minimum width=1cm, minimum height=0.5cm, line width=0.3mm, align=center, fill=green!10] (clslayN) at (-1.2,-1.7) {E\textsubscript{Tok1}};

        \node[block, minimum width=1cm, minimum height=0.5cm, line width=0.3mm, align=center, fill=green!10] (clslayN) at (2.3,-1.7) {E\textsubscript{TokT}};

        \node[block, minimum width=1cm, minimum height=0.5cm, line width=0.3mm, align=center, fill=green!10] (clslayN) at (-2.3,-3.3) {E\textsubscript{[CLS]}};

        \node[block, minimum width=1cm, minimum height=0.5cm, line width=0.3mm, align=center, fill=green!10] (clslayN) at (-1.2,-3.3) {E\textsubscript{Tok1}};

        \node[block, minimum width=1cm, minimum height=0.5cm, line width=0.3mm, align=center, fill=green!10] (clslayN) at (2.3,-3.3) {E\textsubscript{TokT}};
		
		\node[block, minimum width=1cm, minimum height=0.5cm, line width=0.3mm, align=center, fill=yellow!10] (cls) at (-2.3,-4.2) {[CLS]};
		\node[block, minimum width=1cm, minimum height=0.5cm, line width=0.3mm, align=center, fill=yellow!10] (tok1) at (-1.15,-4.2) {Tok1};
		\node[block, minimum width=1cm, minimum height=0.5cm, line width=0.3mm, align=center, fill=yellow!10] (tok2) at (0,-4.2) {Tok2};
		\node[minimum width=1cm, minimum height=0.5cm, line width=0.3mm, align=center] (dots) at (1.15,-4.2) {\dots};
		\node[block, minimum width=1cm, minimum height=0.5cm, line width=0.3mm, align=center, fill=yellow!10] (tokn) at (2.3,-4.2) {TokT};

        \draw[->, line width=0.3mm] (-2.3,-1.4) -- (-2.3,-0.86);
		
        \draw[->, line width=0.3mm] (2.3,-2.4) -- (2.3, -1.95);
		\draw[->, line width=0.3mm] (0,-2.4) -- (0, -1.95);
		\draw[->, line width=0.3mm] (-1.15,-2.4) -- (-1.15, -1.95);
		\draw[->, line width=0.3mm] (-2.3,-2.4) -- (-2.3,-1.95);

		\draw[->, line width=0.3mm] (2.3,-3.05) -- (2.3, -2.6);
		\draw[->, line width=0.3mm] (0,-3.05) -- (0, -2.6);
		\draw[->, line width=0.3mm] (-1.15,-3.05) -- (-1.15, -2.6);
		\draw[->, line width=0.3mm] (-2.3,-3.05) -- (-2.3,-2.6);

		\draw[->, line width=0.3mm] (2.3,-3.95) -- (2.3, -3.5);
		\draw[->, line width=0.3mm] (0,-3.95) -- (0, -3.5);
		\draw[->, line width=0.3mm] (-1.15,-3.95) -- (-1.15, -3.5);
		\draw[->, line width=0.3mm] (-2.3,-3.95) -- (-2.3,-3.5);

		\draw[->, line width=0.3mm] (-2.3,-0.29) -- (-2.3,0.25);
	
		\draw[-, line width=0.3mm] (-3.2,-1.7) -- (-3,-1.7);
        \draw[-, line width=0.3mm] (-3.2,-2.1) -- (-3,-2.1);

        \draw[decorate,decoration={brace,amplitude=4pt},xshift=-2pt,yshift=0pt](-3.2,-2.5) -- (-3.2,-1.7) node [black,midway,xshift=-0.3cm]{\footnotesize $k$};

		\draw[->, line width=0.3mm] (-3.2,-2.5) |- (-3, -0.65);
		\draw[-, line width=0.3mm] (-3.2,-2.5) -- (-3,-2.5);
		
	\end{tikzpicture}
    \caption{\textbf{$\text{Max}_\text{CLS}$.} A max-pooling operation is applied on the \texttt{[CLS]} tokens of the last $k$ layers before classification.}
    \label{fig:maxcls}
\end{subfigure}
\begin{subfigure}[t]{0.48\textwidth}
    \centering
	\begin{tikzpicture}[block/.style={rectangle, draw, line width=0.3mm, rounded corners, font=\ttfamily\scriptsize}]
		\node[block, minimum width=6.0cm, minimum height=0.9cm, line width=0.6mm, rounded corners=5mm, fill=cyan!10, drop shadow] (layer1) at (0,-3.3) {Layer 1}; 
		
		\node[minimum width=6.0cm, minimum height=0.3cm] (dots2) at (0,-2.5) {\dots};
		
		\node[block, minimum width=6.0cm, minimum height=0.9cm, line width=0.6mm, rounded corners=5mm, fill=cyan!10, drop shadow] (layern) at (0,-1.7) {Layer N}; 
		
		\node[minimum width=6.0cm, minimum height=0.3cm] (dots3) at (0,-0.6) {};

        \node[block, minimum width=6.0cm, minimum height=0.9cm, line width=0.6mm, rounded corners=5mm, fill=magenta!10, drop shadow] (attention) at (0,-0.6) {MHA Layer};
		
		\node[block, minimum width=6.0cm, minimum height=0.9cm, line width=0.6mm, rounded corners=5mm, fill=magenta!10, drop shadow] (clshead) at (0,0.5) {Classification Head};

        \node[block, minimum width=1cm, minimum height=0.5cm, line width=0.3mm, align=center, fill=green!10] (clslayN) at (-2.3,-0.6) {E\textsubscript{[CLS]}};

        \node[block, minimum width=1cm, minimum height=0.5cm, line width=0.3mm, align=center, fill=green!10] (clslayN) at (-2.3,-1.7) {E\textsubscript{[CLS]}};

        \node[block, minimum width=1cm, minimum height=0.5cm, line width=0.3mm, align=center, fill=green!10] (clslayN) at (-1.2,-1.7) {E\textsubscript{Tok1}};

        \node[block, minimum width=1cm, minimum height=0.5cm, line width=0.3mm, align=center, fill=green!10] (clslayN) at (2.3,-1.7) {E\textsubscript{TokT}};

        \node[block, minimum width=1cm, minimum height=0.5cm, line width=0.3mm, align=center, fill=green!10] (clslayN) at (-2.3,-3.3) {E\textsubscript{[CLS]}};

        \node[block, minimum width=1cm, minimum height=0.5cm, line width=0.3mm, align=center, fill=green!10] (clslayN) at (-1.2,-3.3) {E\textsubscript{Tok1}};

        \node[block, minimum width=1cm, minimum height=0.5cm, line width=0.3mm, align=center, fill=green!10] (clslayN) at (2.3,-3.3) {E\textsubscript{TokT}};
		
		\node[block, minimum width=1cm, minimum height=0.5cm, line width=0.3mm, align=center, fill=yellow!10] (cls) at (-2.3,-4.2) {[CLS]};
		\node[block, minimum width=1cm, minimum height=0.5cm, line width=0.3mm, align=center, fill=yellow!10] (tok1) at (-1.15,-4.2) {Tok1};
		\node[block, minimum width=1cm, minimum height=0.5cm, line width=0.3mm, align=center, fill=yellow!10] (tok2) at (0,-4.2) {Tok2};
		\node[minimum width=1cm, minimum height=0.5cm, line width=0.3mm, align=center] (dots) at (1.15,-4.2) {\dots};
		\node[block, minimum width=1cm, minimum height=0.5cm, line width=0.3mm, align=center, fill=yellow!10] (tokn) at (2.3,-4.2) {TokT};
		
        \draw[->, line width=0.3mm] (2.3,-2.4) -- (2.3, -1.95);
		\draw[->, line width=0.3mm] (0,-2.4) -- (0, -1.95);
		\draw[->, line width=0.3mm] (-1.15,-2.4) -- (-1.15, -1.95);
		\draw[->, line width=0.3mm] (-2.3,-2.4) -- (-2.3,-1.95);

		\draw[->, line width=0.3mm] (2.3,-3.05) -- (2.3, -2.6);
		\draw[->, line width=0.3mm] (0,-3.05) -- (0, -2.6);
		\draw[->, line width=0.3mm] (-1.15,-3.05) -- (-1.15, -2.6);
		\draw[->, line width=0.3mm] (-2.3,-3.05) -- (-2.3,-2.6);

		\draw[->, line width=0.3mm] (2.3,-3.95) -- (2.3, -3.5);
		\draw[->, line width=0.3mm] (0,-3.95) -- (0, -3.5);
		\draw[->, line width=0.3mm] (-1.15,-3.95) -- (-1.15, -3.5);
		\draw[->, line width=0.3mm] (-2.3,-3.95) -- (-2.3,-3.5);

		\draw[->, line width=0.3mm] (-2.3,-0.29) -- (-2.3,0.25);

        \draw[->, line width=0.3mm] (0,-1.41) -- (0,-0.87);
		\draw[->, line width=0.3mm] (-1.15,-1.41) -- (-1.15,-0.87);
		\draw[->, line width=0.3mm] (-2.3,-1.41) -- (-2.3,-0.87);
		\draw[->, line width=0.3mm] (2.3,-1.41) -- (2.3,-0.87);
		
	\end{tikzpicture}
    \caption{\textbf{MHA.} An additional multi-head attention layer allows the \texttt{[CLS]} token to attend to all tokens of the last layer.}
    \label{fig:mha}
\end{subfigure}\quad
\begin{subfigure}[t]{0.48\textwidth}
    \centering
	\begin{tikzpicture}[block/.style={rectangle, draw, line width=0.3mm, rounded corners, font=\ttfamily\scriptsize}]
		\node[block, minimum width=6.0cm, minimum height=0.9cm, line width=0.6mm, rounded corners=5mm, fill=cyan!10, drop shadow] (layer1) at (0,-3.3) {Layer 1}; 
		
		\node[minimum width=6.0cm, minimum height=0.3cm] (dots2) at (0,-2.5) {\dots};
		
		\node[block, minimum width=6.0cm, minimum height=0.9cm, line width=0.6mm, rounded corners=5mm, fill=cyan!10, drop shadow] (layern) at (0,-1.7) {Layer N}; 
		
		\node[minimum width=6.0cm, minimum height=0.3cm] (dots3) at (0,-0.6) {};
		
		\node[block, minimum width=6.0cm, minimum height=0.9cm, line width=0.6mm, rounded corners=5mm, fill=magenta!10, drop shadow] (attention) at (0,-0.6) {Max};

        \node[block, minimum width=6.0cm, minimum height=0.9cm, line width=0.6mm, rounded corners=5mm, fill=magenta!10, drop shadow] (attention) at (0,0.5) {MHA Layer};
		
		\node[block, minimum width=6.0cm, minimum height=0.9cm, line width=0.6mm, rounded corners=5mm, fill=magenta!10, drop shadow] (clshead) at (0,1.6) {Classification Head};

        \node[block, minimum width=1cm, minimum height=0.5cm, line width=0.3mm, align=center, fill=green!10] (clslayN) at (-2.3,0.5) {E\textsubscript{[CLS]}};

        \node[block, minimum width=1cm, minimum height=0.5cm, line width=0.3mm, align=center, fill=green!10] (clslayN) at (-2.3,-1.7) {E\textsubscript{[CLS]}};

        \node[block, minimum width=1cm, minimum height=0.5cm, line width=0.3mm, align=center, fill=green!10] (clslayN) at (-1.2,-1.7) {E\textsubscript{Tok1}};

        \node[block, minimum width=1cm, minimum height=0.5cm, line width=0.3mm, align=center, fill=green!10] (clslayN) at (2.3,-1.7) {E\textsubscript{TokT}};

        \node[block, minimum width=1cm, minimum height=0.5cm, line width=0.3mm, align=center, fill=green!10] (clslayN) at (-2.3,-3.3) {E\textsubscript{[CLS]}};

        \node[block, minimum width=1cm, minimum height=0.5cm, line width=0.3mm, align=center, fill=green!10] (clslayN) at (-1.2,-3.3) {E\textsubscript{Tok1}};

        \node[block, minimum width=1cm, minimum height=0.5cm, line width=0.3mm, align=center, fill=green!10] (clslayN) at (2.3,-3.3) {E\textsubscript{TokT}};
		
		\node[block, minimum width=1cm, minimum height=0.5cm, line width=0.3mm, align=center, fill=yellow!10] (cls) at (-2.3,-4.2) {[CLS]};
		\node[block, minimum width=1cm, minimum height=0.5cm, line width=0.3mm, align=center, fill=yellow!10] (tok1) at (-1.15,-4.2) {Tok1};
		\node[block, minimum width=1cm, minimum height=0.5cm, line width=0.3mm, align=center, fill=yellow!10] (tok2) at (0,-4.2) {Tok2};
		\node[minimum width=1cm, minimum height=0.5cm, line width=0.3mm, align=center] (dots) at (1.15,-4.2) {\dots};
		\node[block, minimum width=1cm, minimum height=0.5cm, line width=0.3mm, align=center, fill=yellow!10] (tokn) at (2.3,-4.2) {TokT};

        \draw[->, line width=0.3mm] (0,-1.41) -- (0,-0.87);
		\draw[->, line width=0.3mm] (-1.15,-1.41) -- (-1.15,-0.87);
		\draw[->, line width=0.3mm] (-2.3,-1.41) -- (-2.3,-0.87);
		\draw[->, line width=0.3mm] (2.3,-1.41) -- (2.3,-0.87);
		
        \draw[->, line width=0.3mm] (2.3,-2.4) -- (2.3, -1.95);
		\draw[->, line width=0.3mm] (0,-2.4) -- (0, -1.95);
		\draw[->, line width=0.3mm] (-1.15,-2.4) -- (-1.15, -1.95);
		\draw[->, line width=0.3mm] (-2.3,-2.4) -- (-2.3,-1.95);

		\draw[->, line width=0.3mm] (2.3,-3.05) -- (2.3, -2.6);
		\draw[->, line width=0.3mm] (0,-3.05) -- (0, -2.6);
		\draw[->, line width=0.3mm] (-1.15,-3.05) -- (-1.15, -2.6);
		\draw[->, line width=0.3mm] (-2.3,-3.05) -- (-2.3,-2.6);

		\draw[->, line width=0.3mm] (2.3,-3.95) -- (2.3, -3.5);
		\draw[->, line width=0.3mm] (0,-3.95) -- (0, -3.5);
		\draw[->, line width=0.3mm] (-1.15,-3.95) -- (-1.15, -3.5);
		\draw[->, line width=0.3mm] (-2.3,-3.95) -- (-2.3,-3.5);

		\draw[->, line width=0.3mm] (0,-0.29) -- (0, 0.25);
		\draw[->, line width=0.3mm] (-1.15,-0.29) -- (-1.15, 0.25);
		\draw[->, line width=0.3mm] (-2.3,-0.29) -- (-2.3,0.25);
		\draw[->, line width=0.3mm] (2.3,-0.29) -- (2.3,0.25);

		\draw[->, line width=0.3mm] (-2.3,0.8) -- (-2.3, 1.34); 

		\draw[-, line width=0.3mm] (-3.2,-1.7) -- (-3,-1.7);
        \draw[-, line width=0.3mm] (-3.2,-2.1) -- (-3,-2.1);

        \draw[decorate,decoration={brace,amplitude=4pt},xshift=-2pt,yshift=0pt](-3.2,-2.5) -- (-3.2,-1.7) node [black,midway,xshift=-0.3cm]{\footnotesize $k$};

		\draw[->, line width=0.3mm] (-3.2,-2.5) |- (-3, -0.65);
		\draw[-, line width=0.3mm] (-3.2,-2.5) -- (-3,-2.5);
		
	\end{tikzpicture}
    \caption{\textbf{$\text{Max}_\text{Seq}$ + MHA.} A max-pooling operation on the whole sequence is combined with an additional MHA layer.}
    \label{fig:maxbert}
\end{subfigure}
\caption{\textbf{Comparison of four BERT architectures for sequence classification.} \textit{(Left above)} Classical BERT for sequence classification architecture. \textit{(Right above)} Applying max-pooling on the token embeddings of the \texttt{[CLS]} token over the last $k$ layers. \textit{(Left below)} Adding an additional MHA layer before classification. \textit{(Right below)} MaxPoolBERT architecture: After the Nth layer (N = 12 for BERT base), we apply a sequence-wide max-pooling operation over the last $k$ layers (we used $k=3$). The \texttt{[CLS]} token can then attend to every token after the max-pooling and the resulting \texttt{[CLS]} token embedding is used for classification.}
\label{fig:architecures}
\end{figure*}

\section{Refining the \texttt{[CLS]} Token}
\label{sec:refining}

It has been shown that other token representations in the layers of BERT also capture sentence-level representations~\citep{rogers-etal-2020-primer}. We investigate whether the informativeness of the \texttt{[CLS]} token embedding can be further enhanced during fine-tuning, to improve downstream classification results. To do this, we include more \emph{depth} information from other BERT layers and also more \emph{width} information from other tokens within the sequence.
We study different versions of fine-tuning BERT for sequence classification tasks. All variants are described below.
    
\subsection{Preliminaries} 
\paragraph{Final-Layer \texttt{[CLS]} Representation.} As a baseline we use the \texttt{[CLS]} token of the final encoder layer of a fine-tuned vanilla BERT base model~\citep{devlin_etal_2019_bert} for classification (see Figure~\ref{fig:bert}).  Recall that a single layer of BERT can be written as
\begin{equation}
    f_i : \mathbb{R}^{T\times d} \rightarrow \mathbb{R}^{T\times d},
\end{equation}
where $i$ indicates the layer number (BERT base has 12 layers), $T$ is the number of tokens, and $d$ is the dimensionality of each token vector.  
We denote the values of the intermediate layers by $y^{(i)}$:
\begin{equation}
    y^{(1)}=f_1(x), \quad\quad y^{(i+1)} = f_{i+1}(y^{(i)}).
\end{equation}
The classification token of each layer is the first token, i.e., for a sequence of tokens $y^{(i)} = [t_{1i}, ..., t_{Ti}]$ in the $i$th layer,
\begin{equation}
    \texttt{[CLS]}_i = t_{1i} \in\mathbb{R}^{1\times d}.
\end{equation}
The embedding of the \texttt{[CLS]} token serves as the input for the classification head $c$, which we have choosen to be a simple linear layer without an activation function, since we are just interested in the plain expressiveness of the refinement (instead of adding tanh as in the original BERT implementation):
\begin{equation}
    c: \mathbb{R}^{1\times d} \rightarrow \mathbb{R}.
\end{equation}
Thus, the baseline model for sequence classification can be written as:
\begin{equation}
    (c \circ \text{CLS} \circ f_{12} \circ \cdots \circ f_1): \mathbb{R}^{T\times d}  \rightarrow \mathbb{R}
\end{equation}
for BERT base with 12 layers.

\paragraph{Max-pooling operation.}
\label{subsec:maxpool}
The final layers of a BERT model are known to contain the task-specific information. 
In order to utilize not only the last layer but several layers, we have to define a flexible maximizing operation, that can work with several sequences of vectors.  For this, we write $\Theta_{t}^{(k)}\in\mathbb{R}^{k\times t\times d}$ for the tensor that contains the first $t$ token vectors (each $d$ dimensional) of the last $k$ layers.  For instance, $\Theta_{1}^{(1)}\in\mathbb{R}^{1\times 1\times d}$ is the \texttt{[CLS]} token, and $\Theta_{1}^{(k)}\in\mathbb{R}^{k\times 1\times d}$ collects the token vectors the \texttt{[CLS]} token of the last $k$ layers.  Similarly, $\Theta_{T}^{(1)}\in\mathbb{R}^{1\times T\times d}$ contains all token vectors of the last layer, and $\Theta_{T}^{(k)}\in\mathbb{R}^{k\times T\times d}$ all token vectors of the last $k$ layers.

Next, we define an element-wise max-pooling operation that maximizes over the first dimension, i.e., 
\begin{equation}
    \max : \mathbb{R}^{k\times t\times d} \rightarrow \mathbb{R}^{t \times d}.
\end{equation}
Written as Pytorch\footnote{\url{https://pytorch.org/}} code, the operation is \texttt{torch.max(Theta, dim=1)} for $b$-sized minibatches of shape $b\times k\times t\times d$. 

\paragraph{Mean-pooling operation.} Several studies indicate, that max-pooling seems to be a stable choice to aggregate information into a single sentence representation. In the experimental section (Section~\ref{sec:experiments}),  we also consider mean-pooling to challenge these results. For this, we apply an element-wise mean-pooling operation
\begin{equation}
    \text{mean} : \mathbb{R}^{k\times t\times d} \rightarrow \mathbb{R}^{t \times d}
\end{equation}
on every vector of our $k$ chosen layers (defined analogously as the max-pooling operation).

\subsection{Depth-Wise \texttt{[CLS]} Pooling ($\text{Max}_\text{CLS}$)} 
To use the \emph{vertical} information (i.e., more depth) as one possible improvement for BERT's fine-tuning, we take information from the last $k$ layers (instead of only from the last layer):  we extract the last $k$ \texttt{[CLS]} embeddings $[\texttt{[CLS]}_{12-k+1}, \dots, \texttt{[CLS]}_{12}]$ which corresponds to $\Theta_{1}^{(k)}\in\mathbb{R}^{k\times 1\times d}$ (using the notation of the previous paragraph).  Then we apply the element-wise max-pooling operation on the extracted tokens (see Figure~\ref{fig:maxcls}).

\subsection{Token-Wise Attention via Additional MHA Layer ($\text{MHA}$)}  The orthogonal way to enrich the information in the \texttt{[CLS]} token, is to consider \emph{horizontal} information (i.e., more width, see Figure~\ref{fig:mha}).  For this, we include \emph{all} tokens of the last layer. To obtain a single vector, we employ an additional multi-head attention (MHA) layer on the encoder output, but compute the attention only for the \texttt{[CLS]} token.  We write the MHA as  \citep[see][]{NIPS2017_3f5ee243},
\begin{equation}
    \text{MHA(Q,K,V)} = [\text{head}_1, \dots, \text{head}_h]W_0
\end{equation}
where the heads are defined as 
\begin{equation}
    \text{head}_s = \text{Attention}(QW_s^Q, KW_s^K, VW_s^V).
\end{equation}
Using the standard BERT base model with 12 layers, we have $Q = \texttt{[CLS]}_{12}$ and $K = V = y^{(12)}$.
Through the attention mechanism, the \texttt{[CLS]} token can attend to all other tokens once more before classification. Note that the additional MHA layer is not part of the pre-training process and is added and initialized before the fine-tuning process. We use the default initialization of the Pytorch$^1$ multi-head attention implementation which is a Xavier uniform initialization \citep{pmlr-v9-glorot10a}. For the number of attention heads we choose $h=4$.


\begin{wraptable}{r}{0.45\textwidth}
	\centering
    \footnotesize
	\begin{tabular}{lr}
		\textbf{Parameter} & \textbf{Value} \\
		\hline
        learning rate & 2e-5\\
		epochs & 4\\
		batch size & 32\\
		warmup ratio & 0.1\\
		weight decay & 0.01 \\
        \hline
	\end{tabular}
 \caption{Hyperparameters used for all fine-tuning experiments.}
 \label{tab:params}
\end{wraptable}
\newpage
\subsection{Sequence-Wide Pooling with MHA ($\text{Max}_\text{Seq}$ + MHA \& $\text{Mean}_\text{Seq}$ + MHA )} Finally, we combine the additional \textit{depth} and \textit{width} information of \textbf{$\text{Max}_\text{CLS}$} and \textbf{MHA} by extending the max-pooling operation to the whole sequences of the last $k$ layers by using $\max(\Theta_{T}^{(k)})\in\mathbb{R}^{k\times T\times d}$. We call this setup \textbf{$\text{Max}_\text{Seq}$+ MHA}, since the maximum is now along the whole sequence and the additional MHA layer aggregates the pooled information. We call this approach \textbf{MaxPoolBERT} in the following. 
As a variant, we replaced max pooling with mean pooling. We report the results for mean pooling with an additional MHA layer as \textbf{$\text{Mean}_\text{Seq}$+ MHA}. 


\section{Experiments}
\label{sec:experiments}
In order to evaluate each previously presented modification of the BERT architecture for sequence classification, we fine-tune each model on different classification tasks of the GLUE benchmark and compare the results. As a baseline, we use a standard BERT base model \citep{devlin_etal_2019_bert}. In addition, we assess the generalizability of our approach by applying it to a BERT variant, namely RoBERTa base \citep{liu2019roberta}.

\subsection{Datasets}
The General Language Understading Evaluation (GLUE) benchmark \citep{wang-etal-2018-glue} is a well known benchmark for natural language understanding (NLU) and natural language inference (NLI) tasks. We evaluate on the following 9 tasks:
\begin{itemize}
    \item \textbf{CoLA} (Corpus of Linguistic Acceptability \citep{warstadt-etal-2019-neural}): 10,657 sentences from linguistic publications, annotated for grammatical acceptability (\emph{acceptable} or \emph{unacceptable}).
    \item \textbf{MRPC} (Microsoft Research Paraphrase Corpus \citep{dolan-brockett-2005-automatically}): 5,800 sentence pairs from news source, annotated for paraphrase identification (\emph{equivalent} or \emph{not equivalent}).
    \item \textbf{QNLI} (Question NLI): an NLI dataset derived from the Stanford Question Answering Dataset (SQuAD) \citep{rajpurkar-etal-2016-squad} containing question paragraph pairs. The task is to predict if the question is answered by the given paragraph (\emph{entailment} or \emph{no entailment}).
    \item \textbf{MNLI} (Multi-Genre NLI \citep{williams-etal-2018-broad}): includes 433,000 sentence pairs, annotated with three different indicators for entailment (\emph{neutral}, \emph{contradiction} or \emph{entailment}).  MNLI includes both matched (in-domain) and mismatched (cross-domain) sections.
    \item \textbf{SST-2} (The Stanford Sentiment Treebank \citep{socher-etal-2013-recursive}): 215,154 phrases annotated for sentiment analysis (\emph{positive} or \emph{negative}).
    \item \textbf{STS-B} (Semantic Textual Similarity Benchmark \citep{cer-etal-2017-semeval}): 8,630 sentence pairs annotated with a textual similarity score (\textit{from zero to five}).
    \item \textbf{RTE} (Recognizing Textual Entailment \citep{10.1007/11736790_9}): 5,770 sentence pairs annotated for entailment recognition (\emph{entailment} or \emph{no entailment}).
    \item \textbf{QQP} (Quora Question Pairs): 795,000 pairs of questions from Quora, annotated for semantical similarity (\emph{duplicate} or \emph{no duplicate}).
    \item \textbf{WNLI}  \citep[Winograd NLI][]{10.5555/3031843.3031909}: 852 sentence pairs annotated for textual entailment (\emph{entailment} or \emph{no entailment}).
\end{itemize}

\subsection{Experimental Details}

All experiments were run on a single NVIDIA A100 GPU. We used the Huggingface transformers and dataset libraries\footnote{https://huggingface.co/} to implement and train all of our models. Each model was fine-tuned three times with three different random seeds for four epochs. We report the mean of all runs and use the validation sets of all GLUE tasks for evaluation. Experimenting with different values for $k$ (the number of the considered layers), we found that $k=3$ works best (see Appendix \ref{subsec:choiceofk}).
All others hyperparameters are listed in Table~\ref{tab:params}. 


\begin{table}[t]
	\centering
    \scriptsize
	\begin{tabular}{l|cccccccccccccc}
        \textbf{Model} & \textbf{CoLA} & \multicolumn{2}{c}{\textbf{MRPC}} & \textbf{QNLI} & \multicolumn{2}{c}{\textbf{MNLI}} & \textbf{SST-2} & \textbf{STS-B} & \textbf{RTE} & \multicolumn{2}{c}{\textbf{QQP}} & \textbf{WNLI} \\
        &MCC&Acc.&F1&Acc.&m&mm&Acc.&Sp.&Acc.&Acc.&F1&Acc. \\
        \rowcolor{lightgray}\scriptsize{Train Size}& 8.5k & 3.7k & 3.7k & 105k& 393k&393k &67k& 5.75k & 2.5k& 364k & 364k&634\\
        \hline
		\scriptsize{BERT base} &53.59 & 82.43 & 87.49 & 90.96 & 84.27 & 84.57 & 92.55 & 88.47 & 63.42 & 90.65 & 87.40 & 49.77 \\
        \scriptsize{$\text{Max}_\text{CLS}$}& 55.32 & 83.66 & 88.5 & \textbf{91.15} & 84.22 &  84.55 & 92.62 & \textbf{88.97} & 63.06 & 90.59 & 87.33 & 50.23\\
        \scriptsize{MHA} & \textbf{55.88} & 85.38 &89.51& 90.49 &\textbf{84.37}& \textbf{84.67} & 92.32 & 88.04 & 64.98 & 90.67 & \textbf{87.45} & \textbf{55.4}\\
        \scriptsize{$\text{Max}_\text{Seq}$+MHA} & 55.35 & \textbf{85.95} & \textbf{89.78} & 90.73 & 83.82 & 84.24 & \textbf{92.74} & 88.22 & 66.06 & 90.59 & 87.32 & \textbf{55.4}\\
        \scriptsize{$\text{Mean}_\text{Seq}$+MHA} & 55.10 & 85.62 & 89.66 & 90.86 & 83.78 & 84.2 & 92.51 & 87.91 & \textbf{66.67} & \textbf{90.68} & 87.41 & 54.46 \\
        \hline
        \scriptsize{$\Delta$} & 2.29 & 3.52 & 2.29 & 0.19 & 0.24 & 0.1 & 0.19 & 0.5 & 3.25 & 0.03 & 0.05 & 5.63 \\
	\end{tabular}
\caption{\textbf{Our proposed variants improve the performance over BERT base on GLUE validation tasks (average of 3 seeds).} The size of the training data set is highlighted in gray. We report Matthews correlation coefficient (MCC) for CoLA, accuracies for matched (m) and mismatched results (mm) for MNLI, and Spearman correlation (Sp.) for STS-B. Below we report the improvement from the best performing variant over the baseline as $\Delta$.}
\label{tab:glue_val}
\end{table}


\begin{table}
    \centering
    \scriptsize
    \begin{tabular}{l|ccccccccc}
         \textbf{Model} & \textbf{CoLA} $\downarrow$ & 
        \textbf{MRPC} $\downarrow$& \textbf{QNLI} $\downarrow$& \textbf{MNLI} $\downarrow$& \textbf{SST-2} $\downarrow$& \textbf{STSB} $\downarrow$& \textbf{RTE} $\downarrow$& \textbf{QQP} $\downarrow$& \textbf{WNLI} $\downarrow$\\
        \hline
         \scriptsize{BERT Base} & 6.34e-02 & 2.42e-02 & \textbf{2.08e-03}& \textbf{1.97e-03}&\textbf{1.99e-03}&\textbf{3.2e-03} & \textbf{1.78e-02} & 10.8e-04 & 5.86e-02\\
         \scriptsize{$\text{Max}_\text{CLS}$} & 4.55e-02 & \textbf{2.02e-02}& 3.89e-03 & 2.73e-03 & 3.81e-03 & 3.8e-03 & 1.86e-02 & 9.26e-04 & 4.61e-02\\
         \scriptsize{MHA} & 4.3e-02 & 2.1e-02& 5.69e-03 & 2.43e-03 & 3.63e-03 & 4.99e-03 & 1.86e-02 & 8.09e-04 & 4.61e-02\\
         \scriptsize{$\text{Max}_\text{Seq}$ + MHA} & \textbf{4.22e-02} & 2.18e-02 & 5.11e-03 & 4.45e-03 & 3.64e-03 & 4.63e-03 & 1.96e-02 & 7.87e-04 & 4.31e-02\\
         \scriptsize{$\text{Mean}_\text{Seq}$ + MHA} & \textbf{4.22e-02} & 2.03e-02 & 4.49e-03 & 5.04e-03 & 4.46e-03 & 4.55e-03 & 2.12e-02 & \textbf{7.46e-04}  & \textbf{3.97e-02}\\
         \hline
    \end{tabular}
    \caption{\textbf{Standard deviations for three fine-tuning runs with different random seeds.}}
    \label{tab:variances}
\end{table}

\section{Results}
We report the results for all model variants in each task and analyze fine-tuning stability by measuring the standard deviation between runs with different seeds. 

\subsection{Performance Across GLUE Tasks}
The performance of each of our four variants on the GLUE benchmark tasks is presented in Table~\ref{tab:glue_val}. For each task, at least one variant achieves higher performance than the BERT baseline, indicating that our proposed methods for enriching the \texttt{[CLS]} token representation are effective. However, the magnitude of improvement varies across tasks.

\begin{wraptable}{l}{0.5\textwidth}
    \footnotesize
    \centering
    \begin{tabular}{l|c}
      \textbf{Model} & \textbf{GLUE avg.} \\
       \hline
       BERT Base  & 79.63\\
       $\text{Max}_\text{CLS}$ & 80.02\\
       MHA & 80.76\\
       $\text{Max}_\text{Seq}$+MHA& \textbf{80.85}\\
       $\text{Mean}_\text{Seq}$+MHA& 80.75\\
       \hline
       $\Delta$ & 1.25\\
    \end{tabular}
    \caption{\textbf{Average performance across all GLUE tasks.}
MaxPoolBERT shows a consistent gain over BERT base.}
    \label{tab:glue_average}
\end{wraptable}

The \textbf{$\text{Max}_\text{CLS}$} variant, which applies max-pooling over the \texttt{[CLS]} token representations from the last $k$ layers, results in marginal to no improvement for most tasks. Notably, this variant achieves the best performance among all variants on QNLI and STS-B, suggesting that layer-wise max-pooling can be beneficial for certain task types. Both tasks incorporate semantic matching between two texts, thus 
both require nuanced understanding of sentence meaning. 

The \textbf{MHA} variant introduces an additional MHA layer, allowing the final-layer \texttt{[CLS]} token to attend to the full sequence before classification. This variant consistently improves upon the baseline BERT model, indicating that this extra attention step, effectively enhances the model's ability to integrate global context. The biggest improvement is observed on the WNLI dataset, which has the fewest training examples in the GLUE benchmark (634 training examples in total), suggesting that the added attention is particularly helpful in low-resource settings.

The \textbf{$\text{Max}_\text{Seq}$+MHA} variant combines token-wise max-pooling over the sequence with the additional MHA layer. This configuration shows the most consistent improvements, achieving a higher performance than the baseline in $7$ out of $9$ tasks. As shown in Figure~\ref{fig:deltas}, the largest improvements are again seen on datasets with limited training data, such as CoLA, MRPC, RTE and WNLI. These findings suggest that combining sequence-level pooling with attention further enhances robustness in low resource settings. 

Overall, \textbf{$\text{Mean}_\text{Seq}$+MHA} performs similarly well as max-pooling, on RTE it even achieves higher performance than max-pooling. In the end, max-pooling seems to be a better choice as mean-pooling as it works better for most GLUE tasks.

\begin{wraptable}{l}{0.4\textwidth}
    \footnotesize
    \centering
    \begin{tabular}{l|c}
      \textbf{Model} & \textbf{GLUE avg.} \\
       \hline
       RoBERTa Base  & \textbf{82.62}\\
       $\text{Max}_\text{Seq}$+MHA& 82.6\\
       \hline
       $\Delta$ & -0.02\\
    \end{tabular}
    \caption{\textbf{Average performance across all GLUE tasks.}
MaxPoolBERT shows a gain over RoBERTa base.}
    \label{tab:glue_average_roberta}
\end{wraptable}

For clarity, Table~\ref{tab:glue_average} shows the average performance of each model variant across all GLUE tasks. The \textbf{$\text{Max}_\text{Seq}$+ MHA} variant, which we call \textbf{MaxPoolBERT}, achieves the highest overall average. While the average improvement over the baseline is 1.25 points, individual tasks show greater improvements.

\begin{figure}[t]
    \centering
    \begin{subfigure}[t]{0.32\textwidth}
        \includegraphics[width=\textwidth]{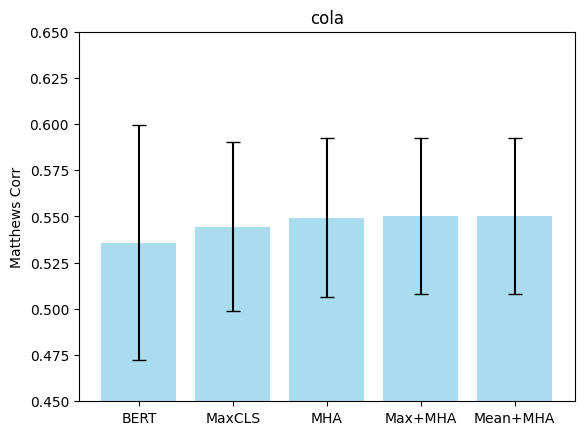}
    \end{subfigure}
    \begin{subfigure}[t]{0.32\textwidth}
        \includegraphics[width=\textwidth]{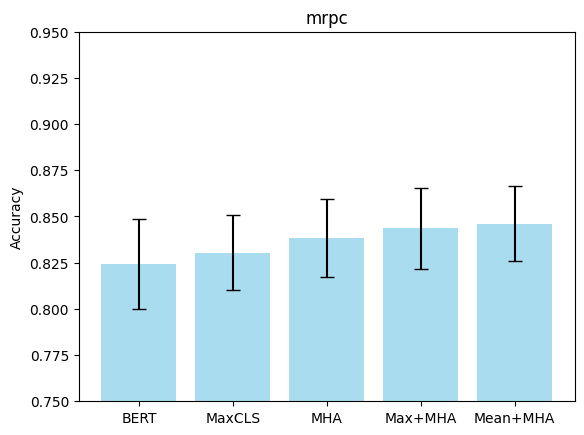}
    \end{subfigure}
    \begin{subfigure}[t]{0.32\textwidth}
        \includegraphics[width=\textwidth]{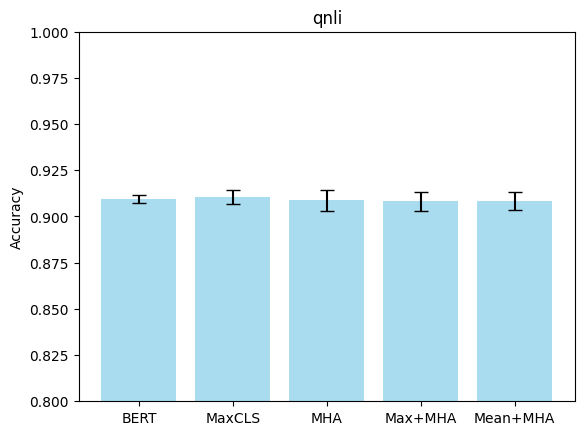}
    \end{subfigure}
    \begin{subfigure}[t]{0.32\textwidth}
        \includegraphics[width=\textwidth]{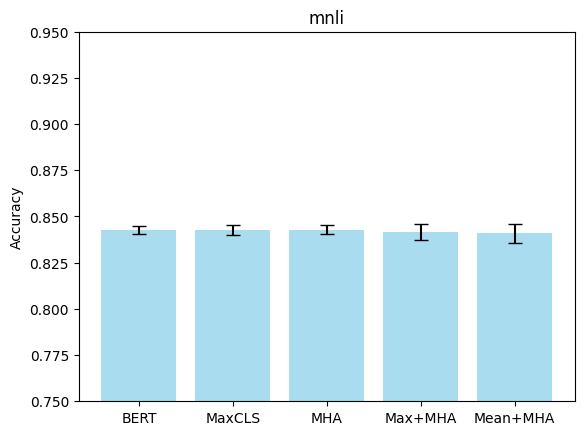}
    \end{subfigure}
    \begin{subfigure}[t]{0.32\textwidth}
        \includegraphics[width=\textwidth]{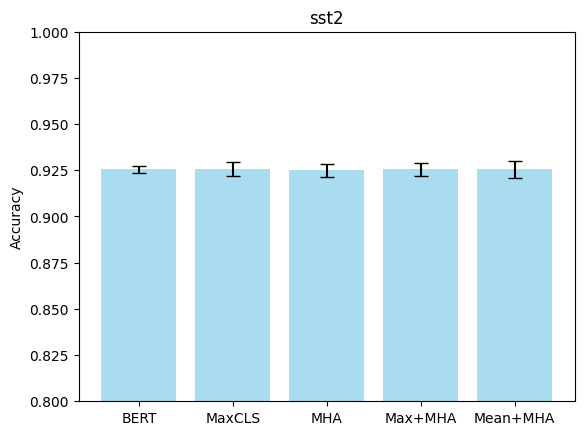}
    \end{subfigure}
    \begin{subfigure}[t]{0.32\textwidth}
        \includegraphics[width=\textwidth]{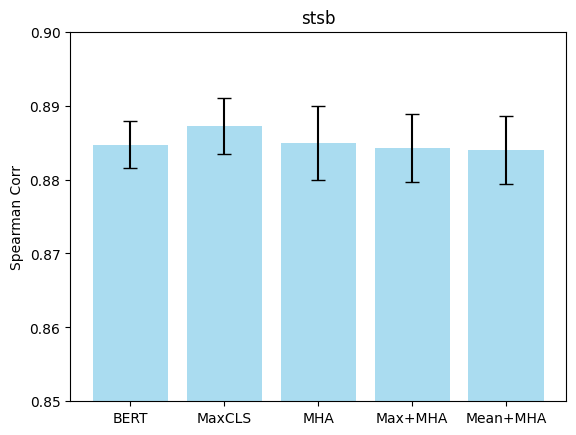}
    \end{subfigure}
    \begin{subfigure}[t]{0.32\textwidth}
        \includegraphics[width=\textwidth]{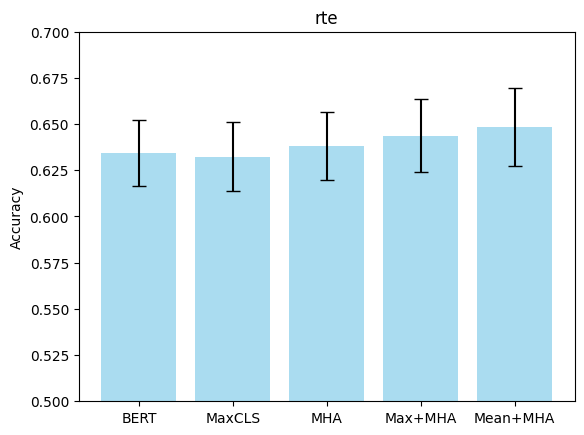}
    \end{subfigure}
    \begin{subfigure}[t]{0.32\textwidth}
        \includegraphics[width=\textwidth]{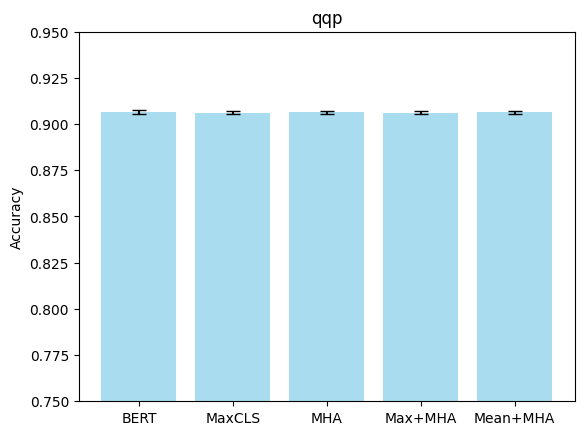}
    \end{subfigure}
    \begin{subfigure}[t]{0.32\textwidth}
        \includegraphics[width=\textwidth]{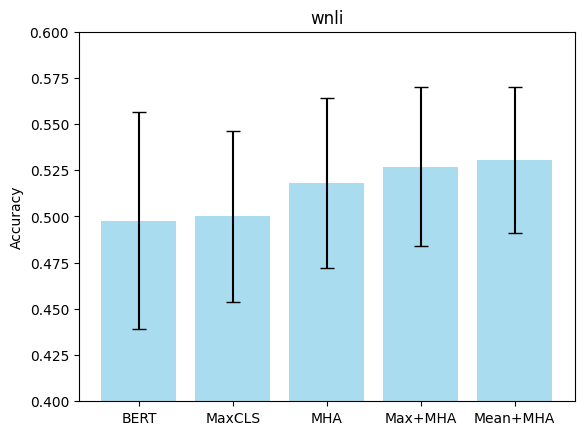}
    \end{subfigure}
    \caption{\textbf{Accuracies for the GLUE benchmark with error bars.}  We show the standard deviation between three fine-tuning runs with three random seeds. Note that the y-axis is shifted but scaled equally across tasks.}
    \label{fig:variances}
\end{figure}

Additionally we apply the max-pooling and MHA combination to a BERT variant, namely RoBERTa \citep{liu2019roberta}. The results are depicted in Table \ref{tab:glue_val_roberta}. For RoBERTa, our max-pooling + MHA variant shows improvements on 4 out of 9 tasks but on average both models perform equally (see Table \ref{tab:glue_average_roberta}). RoBERTa is an optimized version of BERT with a similar architecture, however the effects vary on this BERT variant. A significant improvement can only be observed on the CoLA dataset. 

\begin{table}[h]
	\centering
    \scriptsize
	\begin{tabular}{l|cccccccccccccc}
        \textbf{Model} & \textbf{CoLA} & \multicolumn{2}{c}{\textbf{MRPC}} & \textbf{QNLI} & \multicolumn{2}{c}{\textbf{MNLI}} & \textbf{SST-2} & \textbf{STS-B} & \textbf{RTE} & \multicolumn{2}{c}{\textbf{QQP}} & \textbf{WNLI} \\
        &MCC&Acc.&F1&Acc.&m&mm&Acc.&Sp.&Acc.&Acc.&F1&Acc. \\
        \rowcolor{lightgray}\scriptsize{Train Size}& 8.5k & 3.7k & 3.7k & 105k& 393k&393k &67k& 5.75k & 2.5k& 364k & 364k&634\\
        \hline
		\scriptsize{RoBERTa Base} & 54.96 & \textbf{88.56} & \textbf{91.64} & \textbf{92.29} & \textbf{87.15} & 86.97 & 93.2 & 89.78 & \textbf{73.29} & 90.84 & 87.64 & \textbf{56.34}\\
        \scriptsize{$\text{Max}_\text{Seq}$+MHA} & \textbf{57.43} & 87.58 & 91.01 & 92.15 & 86.99 & \textbf{87.14} & \textbf{93.43} & \textbf{90.49} & 70.4 & \textbf{90.93} & \textbf{87.85} & 55.87\\
        \hline
        \scriptsize{$\Delta$} & 2.47 & -0.98 & -0.63 & -0.14 & -0.16 & 0.17 & 0.23 & 0.71 & -2.89 & 0.09 & 0.21 & -0.47 \\
	\end{tabular}
\caption{\textbf{Performance on GLUE validation tasks for RoBERTa (average of 3 seeds).}} 
\label{tab:glue_val_roberta}
\end{table}


\subsection{Stability on Low-Resource Tasks}
\label{subsec:stability}
To assess fine-tuning stability, which is usually worse for smaller datasets \citep{devlin_etal_2019_bert, Lee2020Mixout:, dodge2020fine}, we run all experiments with three different  seeds 
for each GLUE task. We report the mean accuracy across runs (for CoLA we report Matthews correlation coefficient, for STS-B we report Spearman rank correlation), and include error bars showing the standard deviation of these three runs (see Figure~\ref{fig:variances} and Table~\ref{tab:variances}).

We observe that the stability in fine-tuning remains comparable across model variants for most datasets. However, improvements are observed for datasets with fewer training samples such as CoLA, MRPC, QQP and WNLI, where our variants exhibit reduced variability between runs. These findings suggest that our proposed modifications improve robustness in the low-sample regime. 

\section{Conclusion}

We introduced MaxPoolBERT, a lightweight yet effective refinement of BERT’s classification pipeline that improves the representational quality of the \texttt{[CLS]} token for the BERT base model. Our method leverages max-pooling across layers and tokens, and introduces a multi-head attention layer that allows the \texttt{[CLS]} token to re-aggregate contextual information before classification. These modifications require no changes to pre-training and add minimal overhead to fine-tuning.

Empirical results on the GLUE benchmark demonstrate that MaxPoolBERT outperforms standard BERT base across most tasks, with especially strong improvements in low-resource settings. This suggests that BERT’s native use of the final-layer \texttt{[CLS]} embedding underutilizes available information and that small architectural additions can enhance generalization without sacrificing efficiency.

\subsection*{Limitations}

While MaxPoolBERT improves downstream performance, several limitations remain:

\begin{itemize}
    \item \textbf{No task-specific tuning.} Our experiments use shared hyperparameters across tasks. Further gains could be possible with task-specific settings for pooling depth, attention heads, or training schedules.
    \item \textbf{Model size and generalization.} Our work focuses on BERT base. We also examined one BERT variant but were not able to demonstrate an advantage of applying max-pooling + MHA for RoBERTa. 
    \item \textbf{Scope of evaluation.} We focus on sentence-level classification tasks in GLUE. The applicability of our approach to other tasks, such as token classification, generation, or cross-lingual transfer, is not yet evaluated.
\end{itemize}
In the future we aim to further investigate how to optimize the fine-tuning of small BERT models. While larger models often yield better performance, smaller models are crucial in real-time or resource-constrained environments. The BERT training paradigm following pre-training and fine-tuning has been predominant for several years and is widely used, so it is important to study whether further improvements can be made  through small changes to this learning paradigm.


\subsubsection*{Acknowledgments}
This research has been funded/supported by the Federal Ministry of Education and Research of Germany and the state of North Rhine-Westphalia as part of the Lamarr Institute for Machine Learning and Artificial Intelligence.

\bibliography{iclr2026_conference}

\begin{thebibliography}{38}
\providecommand{\natexlab}[1]{#1}
\providecommand{\url}[1]{\texttt{#1}}
\expandafter\ifx\csname urlstyle\endcsname\relax
  \providecommand{\doi}[1]{doi: #1}\else
  \providecommand{\doi}{doi: \begingroup \urlstyle{rm}\Url}\fi

\bibitem[Bao et~al.(2021)Bao, Zhang, and Zhao]{bao-etal-2021-span-fine}
Rongzhou Bao, Zhuosheng Zhang, and Hai Zhao.
\newblock Span fine-tuning for pre-trained language models.
\newblock In Marie-Francine Moens, Xuanjing Huang, Lucia Specia, and Scott
  Wen-tau Yih (eds.), \emph{Findings of the Association for Computational
  Linguistics: EMNLP 2021}, pp.\  1970--1979, Punta Cana, Dominican Republic,
  November 2021. Association for Computational Linguistics.
\newblock \doi{10.18653/v1/2021.findings-emnlp.169}.
\newblock URL \url{https://aclanthology.org/2021.findings-emnlp.169/}.

\bibitem[Cer et~al.(2017)Cer, Diab, Agirre, Lopez-Gazpio, and
  Specia]{cer-etal-2017-semeval}
Daniel Cer, Mona Diab, Eneko Agirre, I{\~n}igo Lopez-Gazpio, and Lucia Specia.
\newblock {S}em{E}val-2017 task 1: Semantic textual similarity multilingual and
  crosslingual focused evaluation.
\newblock In Steven Bethard, Marine Carpuat, Marianna Apidianaki, Saif~M.
  Mohammad, Daniel Cer, and David Jurgens (eds.), \emph{Proceedings of the 11th
  International Workshop on Semantic Evaluation ({S}em{E}val-2017)}, pp.\
  1--14, Vancouver, Canada, August 2017. Association for Computational
  Linguistics.
\newblock \doi{10.18653/v1/S17-2001}.
\newblock URL \url{https://aclanthology.org/S17-2001/}.

\bibitem[Chang et~al.(2023)Chang, Sun, Ricci, and
  McCallum]{chang-etal-2023-multi}
Haw-Shiuan Chang, Ruei-Yao Sun, Kathryn Ricci, and Andrew McCallum.
\newblock Multi-{CLS} {BERT}: An efficient alternative to traditional
  ensembling.
\newblock In Anna Rogers, Jordan Boyd-Graber, and Naoaki Okazaki (eds.),
  \emph{Proceedings of the 61st Annual Meeting of the Association for
  Computational Linguistics (Volume 1: Long Papers)}, pp.\  821--854, Toronto,
  Canada, July 2023. Association for Computational Linguistics.
\newblock \doi{10.18653/v1/2023.acl-long.48}.
\newblock URL \url{https://aclanthology.org/2023.acl-long.48/}.

\bibitem[Chen et~al.(2023)Chen, Wang, Zhang, Deng, Yukun, and
  Zheng]{chen2023improving}
Qian Chen, Wen Wang, Qinglin Zhang, Chong Deng, Ma~Yukun, and Siqi Zheng.
\newblock Improving bert with hybrid pooling network and drop mask.
\newblock \emph{arXiv preprint arXiv:2307.07258}, 2023.

\bibitem[Dagan et~al.(2006)Dagan, Glickman, and Magnini]{10.1007/11736790_9}
Ido Dagan, Oren Glickman, and Bernardo Magnini.
\newblock The pascal recognising textual entailment challenge.
\newblock In Joaquin Qui{\~{n}}onero-Candela, Ido Dagan, Bernardo Magnini, and
  Florence d'Alch{\'e} Buc (eds.), \emph{Machine Learning Challenges.
  Evaluating Predictive Uncertainty, Visual Object Classification, and
  Recognising Tectual Entailment}, pp.\  177--190, Berlin, Heidelberg, 2006.
  Springer Berlin Heidelberg.
\newblock ISBN 978-3-540-33428-6.

\bibitem[Dao et~al.(2022)Dao, Fu, Ermon, Rudra, and
  R{\'e}]{dao2022flashattention}
Tri Dao, Dan Fu, Stefano Ermon, Atri Rudra, and Christopher R{\'e}.
\newblock Flashattention: Fast and memory-efficient exact attention with
  io-awareness.
\newblock \emph{Advances in Neural Information Processing Systems},
  35:\penalty0 16344--16359, 2022.

\bibitem[Devlin et~al.(2019)Devlin, Chang, Lee, and
  Toutanova]{devlin_etal_2019_bert}
Jacob Devlin, Ming-Wei Chang, Kenton Lee, and Kristina Toutanova.
\newblock {BERT}: Pre-training of deep bidirectional transformers for language
  understanding.
\newblock In \emph{Proceedings of the 2019 Conference of the North {A}merican
  Chapter of the Association for Computational Linguistics: Human Language
  Technologies, Volume 1 (Long and Short Papers)}, pp.\  4171--4186,
  Minneapolis, Minnesota, June 2019. Association for Computational Linguistics.
\newblock \doi{10.18653/v1/N19-1423}.
\newblock URL \url{https://www.aclweb.org/anthology/N19-1423}.

\bibitem[Dodge et~al.(2020)Dodge, Ilharco, Schwartz, Farhadi, Hajishirzi, and
  Smith]{dodge2020fine}
Jesse Dodge, Gabriel Ilharco, Roy Schwartz, Ali Farhadi, Hannaneh Hajishirzi,
  and Noah Smith.
\newblock Fine-tuning pretrained language models: Weight initializations, data
  orders, and early stopping.
\newblock \emph{arXiv preprint arXiv:2002.06305}, 2020.

\bibitem[Dolan \& Brockett(2005)Dolan and
  Brockett]{dolan-brockett-2005-automatically}
William~B. Dolan and Chris Brockett.
\newblock Automatically constructing a corpus of sentential paraphrases.
\newblock In \emph{Proceedings of the Third International Workshop on
  Paraphrasing ({IWP}2005)}, 2005.
\newblock URL \url{https://aclanthology.org/I05-5002/}.

\bibitem[Galal et~al.(2024)Galal, Abdel-Gawad, and Farouk]{galal2024rethinking}
Omar Galal, Ahmed~H Abdel-Gawad, and Mona Farouk.
\newblock Rethinking of bert sentence embedding for text classification.
\newblock \emph{Neural Computing and Applications}, 36\penalty0 (32):\penalty0
  20245--20258, 2024.
\newblock \doi{https://doi.org/10.1007/s00521-024-10212-3}.

\bibitem[Geiping \& Goldstein(2023)Geiping and Goldstein]{pmlr-v202-geiping23a}
Jonas Geiping and Tom Goldstein.
\newblock Cramming: Training a language model on a single {GPU} in one day.
\newblock In Andreas Krause, Emma Brunskill, Kyunghyun Cho, Barbara Engelhardt,
  Sivan Sabato, and Jonathan Scarlett (eds.), \emph{Proceedings of the 40th
  International Conference on Machine Learning}, volume 202 of
  \emph{Proceedings of Machine Learning Research}, pp.\  11117--11143. PMLR,
  23--29 Jul 2023.
\newblock URL \url{https://proceedings.mlr.press/v202/geiping23a.html}.

\bibitem[Glorot \& Bengio(2010)Glorot and Bengio]{pmlr-v9-glorot10a}
Xavier Glorot and Yoshua Bengio.
\newblock Understanding the difficulty of training deep feedforward neural
  networks.
\newblock In Yee~Whye Teh and Mike Titterington (eds.), \emph{Proceedings of
  the Thirteenth International Conference on Artificial Intelligence and
  Statistics}, volume~9 of \emph{Proceedings of Machine Learning Research},
  pp.\  249--256, Chia Laguna Resort, Sardinia, Italy, 13--15 May 2010. PMLR.
\newblock URL \url{https://proceedings.mlr.press/v9/glorot10a.html}.

\bibitem[Goyal et~al.(2020)Goyal, Choudhury, Raje, Chakaravarthy, Sabharwal,
  and Verma]{pmlr-v119-goyal20a}
Saurabh Goyal, Anamitra~Roy Choudhury, Saurabh Raje, Venkatesan Chakaravarthy,
  Yogish Sabharwal, and Ashish Verma.
\newblock {P}o{WER}-{BERT}: Accelerating {BERT} inference via progressive
  word-vector elimination.
\newblock In Hal~Daumé III and Aarti Singh (eds.), \emph{Proceedings of the
  37th International Conference on Machine Learning}, volume 119 of
  \emph{Proceedings of Machine Learning Research}, pp.\  3690--3699. PMLR,
  13--18 Jul 2020.
\newblock URL \url{https://proceedings.mlr.press/v119/goyal20a.html}.

\bibitem[Hao et~al.(2020)Hao, Dong, Wei, and Xu]{hao-etal-2020-investigating}
Yaru Hao, Li~Dong, Furu Wei, and Ke~Xu.
\newblock Investigating learning dynamics of {BERT} fine-tuning.
\newblock In Kam-Fai Wong, Kevin Knight, and Hua Wu (eds.), \emph{Proceedings
  of the 1st Conference of the Asia-Pacific Chapter of the Association for
  Computational Linguistics and the 10th International Joint Conference on
  Natural Language Processing}, pp.\  87--92, Suzhou, China, December 2020.
  Association for Computational Linguistics.
\newblock URL \url{https://aclanthology.org/2020.aacl-main.11}.

\bibitem[Hu et~al.(2024)Hu, Meinel, and Yang]{HU2024101646}
Ting Hu, Christoph Meinel, and Haojin Yang.
\newblock A flexible bert model enabling width- and depth-dynamic inference.
\newblock \emph{Computer Speech \& Language}, 87:\penalty0 101646, 2024.
\newblock ISSN 0885-2308.
\newblock \doi{https://doi.org/10.1016/j.csl.2024.101646}.
\newblock URL
  \url{https://www.sciencedirect.com/science/article/pii/S0885230824000299}.

\bibitem[Hua et~al.(2021)Hua, Li, Dou, Xu, and Luo]{hua-etal-2021-noise}
Hang Hua, Xingjian Li, Dejing Dou, Chengzhong Xu, and Jiebo Luo.
\newblock Noise stability regularization for improving {BERT} fine-tuning.
\newblock In Kristina Toutanova, Anna Rumshisky, Luke Zettlemoyer, Dilek
  Hakkani-Tur, Iz~Beltagy, Steven Bethard, Ryan Cotterell, Tanmoy Chakraborty,
  and Yichao Zhou (eds.), \emph{Proceedings of the 2021 Conference of the North
  American Chapter of the Association for Computational Linguistics: Human
  Language Technologies}, pp.\  3229--3241, Online, June 2021. Association for
  Computational Linguistics.
\newblock \doi{10.18653/v1/2021.naacl-main.258}.
\newblock URL \url{https://aclanthology.org/2021.naacl-main.258}.

\bibitem[Izsak et~al.(2021)Izsak, Berchansky, and Levy]{izsak-etal-2021-train}
Peter Izsak, Moshe Berchansky, and Omer Levy.
\newblock How to train {BERT} with an academic budget.
\newblock In Marie-Francine Moens, Xuanjing Huang, Lucia Specia, and Scott
  Wen-tau Yih (eds.), \emph{Proceedings of the 2021 Conference on Empirical
  Methods in Natural Language Processing}, pp.\  10644--10652, Online and Punta
  Cana, Dominican Republic, November 2021. Association for Computational
  Linguistics.
\newblock \doi{10.18653/v1/2021.emnlp-main.831}.
\newblock URL \url{https://aclanthology.org/2021.emnlp-main.831/}.

\bibitem[Lee et~al.(2020)Lee, Cho, and Kang]{Lee2020Mixout:}
Cheolhyoung Lee, Kyunghyun Cho, and Wanmo Kang.
\newblock Mixout: Effective regularization to finetune large-scale pretrained
  language models.
\newblock In \emph{International Conference on Learning Representations}, 2020.
\newblock URL \url{https://openreview.net/forum?id=HkgaETNtDB}.

\bibitem[Lehe{\v{c}}ka et~al.(2020)Lehe{\v{c}}ka, {\v{S}}vec, Ircing, and
  {\v{S}}m{\'\i}dl]{lehevcka2020adjusting}
Jan Lehe{\v{c}}ka, Jan {\v{S}}vec, Pavel Ircing, and Lubo{\v{s}}
  {\v{S}}m{\'\i}dl.
\newblock Adjusting bert’s pooling layer for large-scale multi-label text
  classification.
\newblock In \emph{International Conference on Text, Speech, and Dialogue},
  pp.\  214--221. Springer, 2020.
\newblock \doi{https://doi.org/10.1007/978-3-030-58323-1_23}.

\bibitem[Levesque et~al.(2012)Levesque, Davis, and
  Morgenstern]{10.5555/3031843.3031909}
Hector~J. Levesque, Ernest Davis, and Leora Morgenstern.
\newblock The winograd schema challenge.
\newblock In \emph{Proceedings of the Thirteenth International Conference on
  Principles of Knowledge Representation and Reasoning}, KR'12, pp.\
  552–561. AAAI Press, 2012.
\newblock ISBN 9781577355601.

\bibitem[Liu et~al.(2019)Liu, Ott, Goyal, Du, Joshi, Chen, Levy, Lewis,
  Zettlemoyer, and Stoyanov]{liu2019roberta}
Yinhan Liu, Myle Ott, Naman Goyal, Jingfei Du, Mandar Joshi, Danqi Chen, Omer
  Levy, Mike Lewis, Luke Zettlemoyer, and Veselin Stoyanov.
\newblock Roberta: A robustly optimized bert pretraining approach.
\newblock \emph{arXiv preprint arXiv:1907.11692}, 2019.
\newblock \doi{https://doi.org/10.48550/arXiv.1907.11692}.

\bibitem[Mosbach et~al.(2021)Mosbach, Andriushchenko, and
  Klakow]{mosbach2021on}
Marius Mosbach, Maksym Andriushchenko, and Dietrich Klakow.
\newblock On the stability of fine-tuning {\{}bert{\}}: Misconceptions,
  explanations, and strong baselines.
\newblock In \emph{International Conference on Learning Representations}, 2021.
\newblock URL \url{https://openreview.net/forum?id=nzpLWnVAyah}.

\bibitem[Nussbaum et~al.(2024)Nussbaum, Morris, Duderstadt, and
  Mulyar]{nussbaum2024nomic}
Zach Nussbaum, John~X Morris, Brandon Duderstadt, and Andriy Mulyar.
\newblock Nomic embed: Training a reproducible long context text embedder.
\newblock \emph{arXiv preprint arXiv:2402.01613}, 2024.

\bibitem[Portes et~al.(2023)Portes, Trott, Havens, KING, Venigalla, Nadeem,
  Sardana, Khudia, and Frankle]{NEURIPS2023_095a6917}
Jacob Portes, Alexander Trott, Sam Havens, DANIEL KING, Abhinav Venigalla, Moin
  Nadeem, Nikhil Sardana, Daya Khudia, and Jonathan Frankle.
\newblock Mosaicbert: A bidirectional encoder optimized for fast pretraining.
\newblock In A.~Oh, T.~Naumann, A.~Globerson, K.~Saenko, M.~Hardt, and
  S.~Levine (eds.), \emph{Advances in Neural Information Processing Systems},
  volume~36, pp.\  3106--3130. Curran Associates, Inc., 2023.
\newblock URL
  \url{https://proceedings.neurips.cc/paper_files/paper/2023/file/095a6917768712b7ccc61acbeecad1d8-Paper-Conference.pdf}.

\bibitem[Rajpurkar et~al.(2016)Rajpurkar, Zhang, Lopyrev, and
  Liang]{rajpurkar-etal-2016-squad}
Pranav Rajpurkar, Jian Zhang, Konstantin Lopyrev, and Percy Liang.
\newblock {SQ}u{AD}: 100,000+ questions for machine comprehension of text.
\newblock In Jian Su, Kevin Duh, and Xavier Carreras (eds.), \emph{Proceedings
  of the 2016 Conference on Empirical Methods in Natural Language Processing},
  pp.\  2383--2392, Austin, Texas, November 2016. Association for Computational
  Linguistics.
\newblock \doi{10.18653/v1/D16-1264}.
\newblock URL \url{https://aclanthology.org/D16-1264/}.

\bibitem[Rogers et~al.(2020)Rogers, Kovaleva, and
  Rumshisky]{rogers-etal-2020-primer}
Anna Rogers, Olga Kovaleva, and Anna Rumshisky.
\newblock A primer in {BERT}ology: What we know about how {BERT} works.
\newblock \emph{Transactions of the Association for Computational Linguistics},
  8:\penalty0 842--866, 2020.
\newblock \doi{10.1162/tacl_a_00349}.
\newblock URL \url{https://aclanthology.org/2020.tacl-1.54/}.

\bibitem[Shazeer(2020)]{shazeer2020glu}
Noam Shazeer.
\newblock Glu variants improve transformer.
\newblock \emph{arXiv preprint arXiv:2002.05202}, 2020.

\bibitem[Socher et~al.(2013)Socher, Perelygin, Wu, Chuang, Manning, Ng, and
  Potts]{socher-etal-2013-recursive}
Richard Socher, Alex Perelygin, Jean Wu, Jason Chuang, Christopher~D. Manning,
  Andrew Ng, and Christopher Potts.
\newblock Recursive deep models for semantic compositionality over a sentiment
  treebank.
\newblock In David Yarowsky, Timothy Baldwin, Anna Korhonen, Karen Livescu, and
  Steven Bethard (eds.), \emph{Proceedings of the 2013 Conference on Empirical
  Methods in Natural Language Processing}, pp.\  1631--1642, Seattle,
  Washington, USA, October 2013. Association for Computational Linguistics.
\newblock URL \url{https://aclanthology.org/D13-1170/}.

\bibitem[Stankevičius \& Lukoševičius(2024)Stankevičius and
  Lukoševičius]{app14198887}
Lukas Stankevičius and Mantas Lukoševičius.
\newblock Extracting sentence embeddings from pretrained transformer models.
\newblock \emph{Applied Sciences}, 14\penalty0 (19), 2024.
\newblock ISSN 2076-3417.
\newblock \doi{10.3390/app14198887}.
\newblock URL \url{https://www.mdpi.com/2076-3417/14/19/8887}.

\bibitem[Su et~al.(2024)Su, Ahmed, Lu, Pan, Bo, and
  Liu]{10.1016/j.neucom.2023.127063}
Jianlin Su, Murtadha Ahmed, Yu~Lu, Shengfeng Pan, Wen Bo, and Yunfeng Liu.
\newblock Roformer: Enhanced transformer with rotary position embedding.
\newblock \emph{Neurocomput.}, 568\penalty0 (C), February 2024.
\newblock ISSN 0925-2312.
\newblock \doi{10.1016/j.neucom.2023.127063}.
\newblock URL \url{https://doi.org/10.1016/j.neucom.2023.127063}.

\bibitem[Toshniwal et~al.(2020)Toshniwal, Shi, Shi, Gao, Livescu, and
  Gimpel]{toshniwal-etal-2020-cross}
Shubham Toshniwal, Haoyue Shi, Bowen Shi, Lingyu Gao, Karen Livescu, and Kevin
  Gimpel.
\newblock A cross-task analysis of text span representations.
\newblock In Spandana Gella, Johannes Welbl, Marek Rei, Fabio Petroni, Patrick
  Lewis, Emma Strubell, Minjoon Seo, and Hannaneh Hajishirzi (eds.),
  \emph{Proceedings of the 5th Workshop on Representation Learning for NLP},
  pp.\  166--176, Online, July 2020. Association for Computational Linguistics.
\newblock \doi{10.18653/v1/2020.repl4nlp-1.20}.
\newblock URL \url{https://aclanthology.org/2020.repl4nlp-1.20/}.

\bibitem[Vaswani et~al.(2017)Vaswani, Shazeer, Parmar, Uszkoreit, Jones, Gomez,
  Kaiser, and Polosukhin]{NIPS2017_3f5ee243}
Ashish Vaswani, Noam Shazeer, Niki Parmar, Jakob Uszkoreit, Llion Jones,
  Aidan~N Gomez, \L~ukasz Kaiser, and Illia Polosukhin.
\newblock Attention is all you need.
\newblock In I.~Guyon, U.~Von Luxburg, S.~Bengio, H.~Wallach, R.~Fergus,
  S.~Vishwanathan, and R.~Garnett (eds.), \emph{Advances in Neural Information
  Processing Systems}, volume~30. Curran Associates, Inc., 2017.
\newblock URL
  \url{https://proceedings.neurips.cc/paper_files/paper/2017/file/3f5ee243547dee91fbd053c1c4a845aa-Paper.pdf}.

\bibitem[Wang et~al.(2018)Wang, Singh, Michael, Hill, Levy, and
  Bowman]{wang-etal-2018-glue}
Alex Wang, Amanpreet Singh, Julian Michael, Felix Hill, Omer Levy, and Samuel
  Bowman.
\newblock {GLUE}: A multi-task benchmark and analysis platform for natural
  language understanding.
\newblock In Tal Linzen, Grzegorz Chrupa{\l}a, and Afra Alishahi (eds.),
  \emph{Proceedings of the 2018 {EMNLP} Workshop {B}lackbox{NLP}: Analyzing and
  Interpreting Neural Networks for {NLP}}, pp.\  353--355, Brussels, Belgium,
  November 2018. Association for Computational Linguistics.
\newblock \doi{10.18653/v1/W18-5446}.
\newblock URL \url{https://aclanthology.org/W18-5446}.

\bibitem[Warner et~al.(2024)Warner, Chaffin, Clavi{\'e}, Weller, Hallstr{\"o}m,
  Taghadouini, Gallagher, Biswas, Ladhak, Aarsen, et~al.]{warner2024smarter}
Benjamin Warner, Antoine Chaffin, Benjamin Clavi{\'e}, Orion Weller, Oskar
  Hallstr{\"o}m, Said Taghadouini, Alexis Gallagher, Raja Biswas, Faisal
  Ladhak, Tom Aarsen, et~al.
\newblock Smarter, better, faster, longer: A modern bidirectional encoder for
  fast, memory efficient, and long context finetuning and inference.
\newblock \emph{arXiv preprint arXiv:2412.13663}, 2024.

\bibitem[Warstadt et~al.(2019)Warstadt, Singh, and
  Bowman]{warstadt-etal-2019-neural}
Alex Warstadt, Amanpreet Singh, and Samuel~R. Bowman.
\newblock Neural network acceptability judgments.
\newblock \emph{Transactions of the Association for Computational Linguistics},
  7:\penalty0 625--641, 2019.
\newblock \doi{10.1162/tacl_a_00290}.
\newblock URL \url{https://aclanthology.org/Q19-1040/}.

\bibitem[Williams et~al.(2018)Williams, Nangia, and
  Bowman]{williams-etal-2018-broad}
Adina Williams, Nikita Nangia, and Samuel Bowman.
\newblock A broad-coverage challenge corpus for sentence understanding through
  inference.
\newblock In Marilyn Walker, Heng Ji, and Amanda Stent (eds.),
  \emph{Proceedings of the 2018 Conference of the North {A}merican Chapter of
  the Association for Computational Linguistics: Human Language Technologies,
  Volume 1 (Long Papers)}, pp.\  1112--1122, New Orleans, Louisiana, June 2018.
  Association for Computational Linguistics.
\newblock \doi{10.18653/v1/N18-1101}.
\newblock URL \url{https://aclanthology.org/N18-1101/}.

\bibitem[Xu et~al.(2023)Xu, Qiu, Zhou, and Huang]{10.1007/s11390-021-1119-0}
Yi-Ge Xu, Xi-Peng Qiu, Li-Gao Zhou, and Xuan-Jing Huang.
\newblock Improving bert fine-tuning via self-ensemble and self-distillation.
\newblock \emph{J. Comput. Sci. Technol.}, 38\penalty0 (4):\penalty0 853–866,
  July 2023.
\newblock ISSN 1000-9000.
\newblock \doi{10.1007/s11390-021-1119-0}.
\newblock URL \url{https://doi.org/10.1007/s11390-021-1119-0}.

\bibitem[Zhang et~al.(2021)Zhang, Wu, Katiyar, Weinberger, and
  Artzi]{zhang2021revisiting}
Tianyi Zhang, Felix Wu, Arzoo Katiyar, Kilian~Q Weinberger, and Yoav Artzi.
\newblock Revisiting few-sample {\{}bert{\}} fine-tuning.
\newblock In \emph{International Conference on Learning Representations}, 2021.
\newblock URL \url{https://openreview.net/forum?id=cO1IH43yUF}.

\end{thebibliography}
\bibliographystyle{iclr2026_conference}

\appendix
\section{Appendix}
\label{sec:appendix}
\subsection{Ablations}
We conduct ablation studies to test different modifications of our architectures. We describe all experiments and report their results in the following.

\subsubsection{Choice of $k$}
\label{subsec:choiceofk}
We experiment with the choice of $k$ for the max-pooling layer on the smaller GLUE datasets (CoLA, MRPC and RTE) and report the results in the following in Table~\ref{tab:glue_k_ablations}. Because we average over three runs with different random seeds, the choice of k does not have an immense influence on performance, but it is apparent that $k = 3$ is the best choice on the data sets tested. 

\begin{wraptable}{r}{0.5\textwidth}
	\centering
    \scriptsize
	\begin{tabular}{l|cccc}
         & \textbf{CoLA} & \multicolumn{2}{c}{\textbf{MRPC}} & \textbf{RTE} \\
        &Acc.&Acc.&F1&Acc.\\
        \hline
        $k = 1$ & 54.85 & 83.58 & 88.44 & 63.18\\
        $k = 2$ & 55.76 & 85.21 & 89.29 & 65.34\\
        $k = 3$ & 55.35 & \textbf{85.95} & \textbf{89.78} & \textbf{66.06} \\
        $k = 4$ & \textbf{56.42} & 85.29 & 89.27 & 65.34 \\
        $k = 6$ & 55.65 & 85.13 & 89.25 & 65.7\\
        $k = 12$ & 55.41 & 85.21 & 89.17 & 65.46 \\
        \hline
	\end{tabular}
\caption{\textbf{Effect of max-pooling depth $k$ on small GLUE tasks.}
$k=3$ generally yields best results.}
\label{tab:glue_k_ablations}
\end{wraptable}

\subsubsection{Time Difference}
To evaluate differences in fine-tuning and inference time, we measured the time to fine-tune both standard BERT and MaxPoolBERT for four epochs on the MRPC dataset on a single A100 GPU. We also measured the inference time on the MRPC validation set for both model variants. Fine-tuning BERT took 289.298 seconds (approx. 72.324 seconds per epoch), inference on the validation set took 2.9935 seconds. In contrast, fine-tuning MaxPoolBERT on the MRPC dataset takes: 294.504 seconds (approx. 73.626 seconds per epoch). Inference time on the validation set is 3.0284 seconds. That is a difference of approximately 1.3 seconds per epoch of fine-tuning and 0.035 seconds difference for inference, which is neglectable. 

\end{document}